\documentclass[10pt,twocolumn,letterpaper]{article}

\usepackage{iccv}
\usepackage{times}
\usepackage{booktabs}
\usepackage{epsfig}
\usepackage{graphicx}
\usepackage{amsmath}
\usepackage{amssymb}
\usepackage{bbm}
\usepackage{bm}
\usepackage{algorithmic}
\usepackage{setspace}
\usepackage{wrapfig}
\usepackage{array}
\usepackage{caption}
\usepackage{amsthm}
\usepackage{xcolor}  
\usepackage{indentfirst}
\usepackage{diagbox}
\usepackage[ruled]{algorithm2e}
\usepackage{comment}
\usepackage{multirow} 
\usepackage{colortbl}
\usepackage{textcomp}
\usepackage{fancyhdr}

\usepackage{subcaption}
\usepackage[pagebackref=true,breaklinks=true,letterpaper=true,colorlinks,bookmarks=false]{hyperref}

\usepackage[capitalize]{cleveref}

\iccvfinalcopy 

\def\iccvPaperID{11378} 

\ificcvfinal\fi

\begin{document}
\title{Enhancing Sample Utilization through Sample Adaptive \\ Augmentation in Semi-Supervised Learning}

\author{Guan Gui{\small $^{1}$}, ~Zhen Zhao{\small $^{2}$}, ~Lei Qi{\small $^{3}$}, ~Luping Zhou{\small $^{2}$}, ~Lei Wang{\small $^{4}$}, ~Yinghuan Shi{\small $^{1}$}\thanks{Corresponding author: Yinghuan Shi. Guan Gui, Yinghuan Shi are with the National Key Laboratory for Novel Software Technology and the National Institute of Healthcare Data Science, Nanjing University. Lei Qi is with the school of Computer Science and Engineering, Southeast University. This Work is supported by NSFC Program (62222604, 62206052, 62192783), China Postdoctoral Science Foundation Project (2023T160100), Jiangsu Natural Science Foundation Project (BK20210224), and CCF-Lenovo Bule Ocean Research Fund.}\\
$^{1}$Nanjing University~~ $^{2}$University of Sydney~~$^{3}$Southeast University~~ $^{4}$University of Wollongong\\
\normalsize
\tt\normalsize guiguan@smail.nju.edu.cn, \{zhen.zhao, luping.zhou\}@sydney.edu.au\\
\tt\normalsize qilei@seu.edu.cn, leiw@uow.edu.au, syh@nju.edu.cn 
}

\maketitle

\begin{abstract}
In semi-supervised learning, unlabeled samples can be utilized through augmentation and consistency regularization. 
However, we observed certain samples, even undergoing strong augmentation, are still correctly classified with high confidence, resulting in a loss close to zero. 
It indicates that these samples have been already learned well and do not provide any additional optimization benefits to the model. We refer to these samples as ``naive samples".
Unfortunately, existing SSL models overlook the characteristics of naive samples, and they just apply the same learning strategy to all samples. 
To further optimize the SSL model, we emphasize the importance of giving attention to naive samples and augmenting them in a more diverse manner.
Sample adaptive augmentation (SAA) is proposed for this stated purpose and consists of two modules: 1) sample selection module; 2) sample augmentation module. 
Specifically, the sample selection module picks out {naive samples} based on historical training information at each epoch, then the naive samples will be augmented in a more diverse manner in the sample augmentation module.
Thanks to the extreme ease of implementation of the above modules, SAA is advantageous for being simple and lightweight. We add SAA on top of FixMatch and FlexMatch respectively, and experiments demonstrate SAA can significantly improve the models.
For example, SAA helped improve the accuracy of FixMatch from 92.50\% to 94.76\% and that of FlexMatch from 95.01\% to 95.31\% on CIFAR-10 with 40 labels.
The code is available at \url{https://github.com/GuanGui-nju/SAA}.
\end{abstract}

\section{Introduction}

For the sake of reducing the cost of manual labeling, semi-supervised learning (SSL), which focuses on how to learn from unlabeled data, is a longstanding yet significant research topic in vision applications. 
Recently, data augmentation techniques and consistency regularization have been proven to be effective ways of utilizing unlabeled data. 
For example, FixMatch~\cite{FixMatch} encourages consistency in predictions between the weakly and strongly augmented versions, and it achieves an accuracy of 92.50\% on the CIFAR-10 task with only 40 labels.

\begin{figure}[t]
  \centering
  \begin{subfigure}{1\linewidth}
  \centering
      \includegraphics[width=0.95\linewidth]{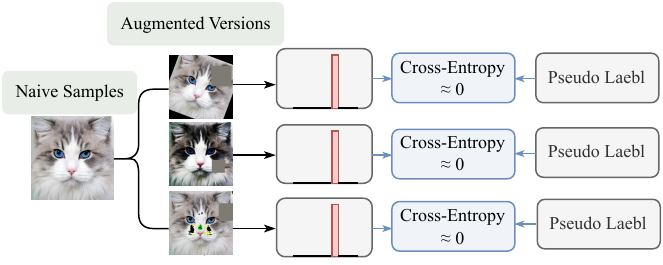}
      \caption{An example of \textit{naive sample}.}
      \label{fig:intro-b}
  \end{subfigure}
\begin{subfigure}{1\linewidth}
\centering
    \includegraphics[width=0.44\linewidth]{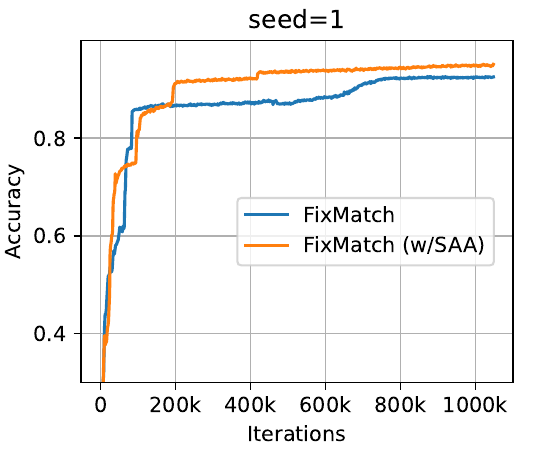}
    \includegraphics[width=0.44\linewidth]{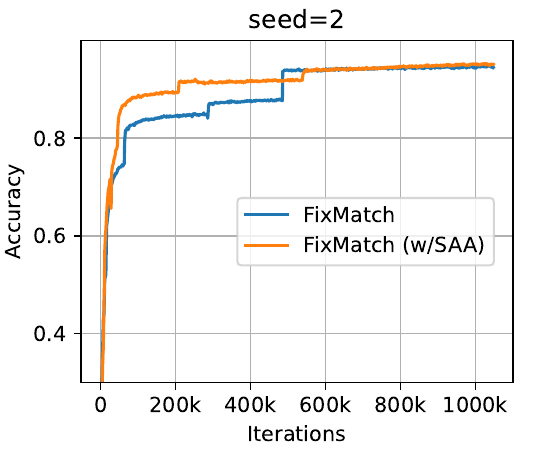}
    \caption{Model performance during training.}
    \label{fig:intro-a}
\end{subfigure}
  \caption{(a) shows an example of \textit{naive sample}. Its augmented versions are correctly classified with high confidence, resulting in the loss close to 0. (b) shows the model performance during FixMatch training. Performance improvements are slow or even stagnant for a period of time.}
  \label{fig:short}
  \vspace{-10pt}
\end{figure}

However, not all unlabeled samples are effectively utilized even with strong augmentation. 
In Figure~\ref{fig:intro-b}, if the strongly augmented versions are correctly classified with high confidence, leading to a loss close to zero, it indicates that the sample has already been learned well and cannot further improve the model's performance.
In other words, the sample was not effectively utilized to benefit model training, and we call this sample ``\textit{naive sample}". When the training process contains a large number of \textit{naive samples}, it can cause slow or even stagnant model performance improvements, as shown in Figure~\ref{fig:intro-a}. 

Unfortunately, existing SSL models~\cite{semireview} overlook the critical point of whether all samples are effectively utilized. Typically, these models apply the same fixed strong augmentation strategy to all samples, resulting in some strongly augmented versions that do not benefit the model train.

We emphasize that the key to alleviating this problem lies in how to further explore the value of the \textit{naive samples} through new learning strategies.
A natural idea that reminds us is to develop sample adaptive augmentation (SAA) to identify \textit{naive samples} and increase their diversity after augmentation. Our proposed SAA is simple yet effective, which consists of two modules: 1) sample selection module and 2) sample augmentation module. The former is responsible for picking out \textit{naive samples} in each epoch, and the latter applies a more diverse augmentation strategy for \textit{naive samples}.
Specifically, in the sample selection module, we first update the historical loss of the samples with exponential moving average (EMA) in each epoch, then these samples will be divided into two parts. The part of the samples with a smaller historical loss is considered to be the \textit{naive sample}. Since historical loss captures the impact of the sample on model training, this approach allows us to identify samples that are not effectively utilized and would benefit from more diverse augmentation.
While in the sample augmentation module, the more diverse augmented version of \textit{naive sample} will be obtained by regrouping multiple strong augmented images, and the remaining samples are applied with the original strong augmentation. 

Our proposed SAA is simple to implement, requiring only a few lines of code to add our proposed modules to the FixMatch or FlexMatch in PyTorch. It is also lightweight in terms of memory and computation, \ie, SAA only needs to add two additional vectors and update them in each epoch, making it an efficient solution for improving SSL models.

We extended FixMatch and FlexMatch with SAA and conducted experiments on SSL benchmarks. The results of the experiments demonstrate that SAA can significantly improve performance. In summary, our contribution can be summarized as follows:
\begin{itemize}
    \item \textbf{We identify ineffectively utilized samples and emphasize that they should be given more attention.} Under the consistency regularization based on data augmentation, some strongly augmented versions are not beneficial to model training, which results in the values of these samples not being fully exploited and makes the model performance slow to improve. We refer to them as \textit{``naive sample"}, and emphasize that they should be learned with a new learning strategy.
    \item  \textbf{We propose SAA to make better use of \textit{naive sample}.} To increase the probability that the augmented versions can benefit the model training, a simple yet effective method, sample adaptive augmentation (SAA), is proposed for identifying the \textit{naive samples} and augmenting them in a more diverse manner.\textit{}
    \item  \textbf{We verify the validity of SAA on SSL benchmarks.} Using FixMatch and FlexMatch as the base framework, we proved that our approach can achieve state-of-the-art performance. For example, on CIFAR-10 with 40 labels, SAA helps FixMatch improve its accuracy from 92.50\% to 94.76\%, and helps FlexMatch improve its accuracy from 95.01\% to 95.31\%.
\end{itemize}

\section{Related Work}

\subsection{Semi-Supervised Learning}
Consistency regularization (CR)~\cite{CR} is the main way to exploit unlabeled data in semi-supervised learning (SSL). The conventional implementation is to perturb the samples and then encourage the model to maintain a consistent prediction. The manner of perturbation has been studied in a variety of ways, \eg, stochastic augmentation and drop out~\cite{PiModel,sup1}, feature perturbations~\cite{FeatMatch}, adversarial perturbations~\cite{virtual}, model perturbations~\cite{MeanTeacher}. ~\cite{MixMatch,ICT,ReMixMatch} apply mixup to blend the images, also a perturbation of the image. With strong augmentation technique~\cite{RandAugment,cutout}, FixMatch applies a consistency regularization between the weakly augmented and strongly augmented versions, allowing the model to learn a greater diversity of images over long iterations. This approach has greatly simplified the framework and has led to breakthroughs in semi-supervised learning milestones. As we have previously analyzed, the framework applies the same fixed augmentation strategy to all images, which results in the \textit{naive samples} not being fully utilized.

Due to the superiority of the FixMatch framework, a large number of SSL works~\cite{FlexMatch,duan2022rda,Dash,simmatch,SCMatch,duan2022mutexmatch,zhao2022dc} are now based on it for further optimization, but none of the work considers the effectiveness of utilization of \textit{naive samples}. 
~\cite{duan2022rda, zhao2022dc} focuses on improving the quality of the pseudo labels by learning the distribution of unlabeled data. 
~\cite{simmatch, SCMatch} focus on learning the similarity relationship between samples or super-classes.
~\cite{duan2022mutexmatch, Dash, FlexMatch} all emphasize the utilization of samples with low confidence.
These works also allow the model to learn more samples within a certain number of iterations to some extent, but still ignore the issue of the validity of the augmentation, resulting in these augmented samples still possibly unhelpful to the training of the mode. 

To the best of our knowledge, none of the SSL works has considered the utilization of \textit{naive samples}.
~\cite{FlexMatch} treats each category differently and adjusts the threshold for each category, but we are considering treating each sample differently, with no relationship to the category.

\subsection{Hard Example Mining}
Our work is somewhat related to hard example mining, with the difference that we focus on \textit{naive samples} that do not benefit model training, while they focus on hard samples that damage model training.
A more common approach used to select hard samples is to rely on loss information between the sample and the ground truth~\cite{hard1,hard9,hard11,hard10}. 
This is related to our approach, but the unlabeled sample is selected by its consistency loss due to the lack of ground-truth. In addition to this, distance metrics~\cite{hard6,hard16} and false positive samples~\cite{hard7,hard12,hard13,hard14} are common methods of hard sample selection. 
However, most of these methods for mining hard samples rely on labels, which is not practical under SSL. Both our proposed method and the field involve the selection of samples, the difference being that they focus on the selection of hard samples that are difficult to train, while we focus on \textit{naive sample} that contributes no information of model training.

\subsection{Date Augmentation}
Data augmentation is an effective way of expanding the data space~\cite{augmentation_survey}, which we roughly classify into the following categories: 1) Single perturbation~\cite{aug1,aug2,cutout}; 2) image blending based~\cite{aug3,aug4,aug5,aug7}; 3) learning based~\cite{aug9,aug10,aug11}; and 4) search based~\cite{aug12,aug13,RandAugment}. Common operations for perturbing a single image include geometric transformations, color transformations, noise injection~\cite{aug1}, random erasing~\cite{aug15}, kernel flters~\cite{aug2}, cutout~\cite{cutout}.  ~\cite{aug3,aug4} direct mixing of the contents of two images, ~\cite{aug5} mixes image patches, and~\cite{aug6,aug7} Mix the content and style. For learning based strategy, adversarial training~\cite{aug9,aug16} and GAN-based~\cite{aug10,aug11} train the network to obtain augmented images. ~\cite{aug12, aug13} find the best combination in the perturbation space using a search strategy. 

However, single perturbation and image blending-based methods are limited to enhancing the diversity of images. For learning and search-based methods, although they yield augmented images that facilitate model training, their time consumption is huge and therefore this is not suitable for training time-consuming CR-based SSL models. ~\cite{RandAugment} combines random transformations to remove the search process, and is favored by the SSL model. ~\cite{cut_once} cuts the image and augments the patches, which enhances image diversity and has been validated to be effective on several tasks. Our augmentation is related to~\cite{cut_once}, but we augmented on the image, not on patches. We will further discuss it in the experimental section.

\begin{figure*}[t]
  \centering
  \begin{subfigure}{0.44\linewidth}
  \centering
    \includegraphics[scale=0.46]{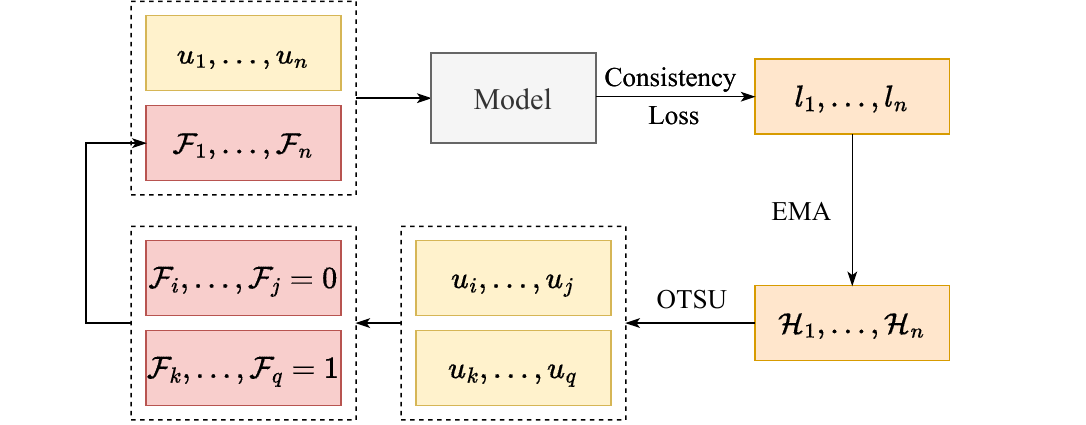}
    \caption{Sample selection module in SAA.}
    \label{fig:short-a}
  \end{subfigure}
  \hfill
  \begin{subfigure}{0.53\linewidth}
  \centering
    \includegraphics[scale=0.8]{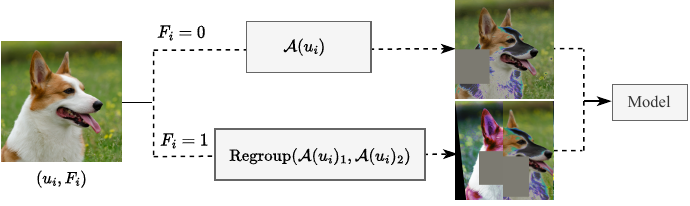}
    \caption{Sample augmentation module in SAA.}
    \label{fig:short-b}
  \end{subfigure}
  \caption{Overview of our method SAA. The core insight of SAA is that dynamically adjusts augmentation for samples, thus allowing \textit{naive samples} to be used more effectively. In detail, SAA consists of two modules: sample selection module and sample augmentation module. (a). 
  Each sample $u_i$ corresponds to a marker $\mathcal{F}_i$ and historical loss $\mathcal{H}_i$. In each epoch, samples' consistency losses are recorded and their historical losses $\mathcal{H}_i$ are updated with EMA. Then based on the historical losses, we divide these samples into two parts by OTSU. 
  The part of the samples with a smaller historical loss are \textit{naive samples}, and their markers are set to 1, then the rest of the markers are set to 0. (b). Sample $u_i$ is augmented in different ways depending on the marker $\mathcal{F}_i$, \ie, if $\mathcal{F}_i=0$, it is strongly augmented once, if $\mathcal{F}_i=1$, it is strongly augmented twice, and the two augmented images are regrouped into one image. The regrouping may be in two parts top-bottom or two parts left-right, which is chosen randomly with a probability of 0.5.}

  \label{fig:short}
\end{figure*}

\section{Preliminary and Background}

\subsection{Problem Setting}
In semi-supervised learning, we denote labeled set $\mathcal{X}=\{(x_1,y_1),(x_2,y_2),\dots,(x_M,y_M)\}$, where $y_i$ is the label of the $i$-th labeled sample $x_i$. We also denote unlabeled set $\mathcal{U}=\{u_1,u_2,\dots,u_N\}$, where $u_i$ denotes $i$-th unlabeled sample, and typically $|\mathcal{X}| \ll |\mathcal{U}|$.
In the implementation, the samples are provided on a per batch basis in each iteration, with a batch of labeled data $\mathcal{X}$ and batches of unlabeled data $\mathcal{U}$. 
How to use this unlabeled data for learning is the focus of SSL research.
Generally, in consistency regularization (CR) based SSL models, unlabeled data generate different versions by perturbation, and then the model is encouraged to be consistent in its representations or predictions of these versions.

\subsection{Preliminary for CR-based SSL Models}
Strong augmentation is a good means of applying consistency regularization, and FixMatch~\cite{FixMatch} is representative of this idea.
Many recent semi-supervised works~\cite{FlexMatch,Dash,CoMatch,simmatch,SCMatch} have also used FixMatch as a basis for further optimization, and to clearly introduce our approach, we also use FixMatch as a base framework. 

We first review FixMatch, whose fundamental idea is that produce pseudo labels on weakly-augmented versions and use them as training targets for their corresponding strongly augmented versions. Of them, the weak augmentation $\mathcal{\alpha}(\cdot)$ includes standard flip and shift operations, while the strong augmentation strategy $\mathcal{A}(\cdot)$ consists of RandAugment~\cite{RandAugment} and CutOut~\cite{cutout}.

Let $p_i^w$ and $p_i^s$ represent the model's prediction on $\mathcal{\alpha}(u_i)$ and $\mathcal{A}(u_i)$, respectively. Then this consistency regularization based unsupervised loss for unlabeled samples is,
\begin{equation}
    \mathcal{L}_{unsup}=\dfrac{1}{|\mathcal{U}|}\sum_{i=1}^{|\mathcal{U}|}{\mathbbm{1}(\max(p^w_i)\ge\tau_c)H(p^w_i,p^s_i)}.
    \label{equ:uloss}
\end{equation}
where $H (p_1, p_2)$ denotes the standard cross entropy between $p_1$ and $p_2$, and $\tau_c$ is a pre-defined threshold to retain only high-confidence pseudo-labels. As discussed in FixMatch~\cite{FixMatch}, $\tau_c$ is commonly set as a large value to alleviate the confirmation bias~\cite{bias2020} in SSL. Let $p_i$ denotes the model's prediction of $\mathcal{\alpha}(x_i)$, then the supervised loss for labeled samples is,
\begin{equation}
    \mathcal{L}_{sup}=\dfrac{1}{|\mathcal{X}|}\sum_{i=1}^{|\mathcal{X}|}H(q_i,y_i).
    \label{equ:xloss}
\end{equation}
Finally, the total losses can be expressed as,
\begin{equation}
    \mathcal{L}=\mathcal{L}_{sup}+\lambda \mathcal{L}_{unsup}.
    \label{equ:tloss}
\end{equation}
where $\lambda$ is the weight of $\mathcal{L}_{unsup}$.

\subsection{Characteristics and Impact of Naive Samples}
\begin{wrapfigure}{r}{3cm}
\vspace{-0.5cm}
\hspace{-0.70cm}
\centering
\includegraphics[scale=0.35]{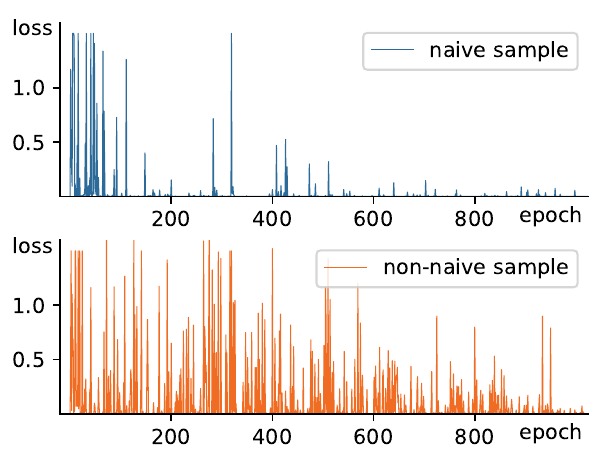}
\captionsetup{font=scriptsize}
\hspace{-0.70cm}
\caption{loss of \textit{non-naive sample} (bottom) and \textit{naive sample} (top).}
\label{fig:intro:twoprop}
\label{fig:loss-1}
\vspace{-0.3cm}
\end{wrapfigure}
Taking FixMatch as an example, we tracked the loss of a \textit{naive sample} and a \textit{non-naive sample} separately, as shown in Figure~\ref{fig:loss-1}. Note that we show the original loss without thresholds to exclude the interference of confidence thresholds on the loss values. It can be observed that in most epochs, the \textit{naive sample's} cross-entropy losses are close to 0, indicating that the learning of $\mathcal{A}(u_i)$ does not contribute to the model's training progress.

With the same augmentation, the \textit{non-naive sample} can encourage model optimization in the long term, whereas \textit{naive samples} cannot. 
This confirms the necessity for dedicated attention to \textit{naive samples} and the development of new augmentation strategies that can better exploit their potential value. 
Moreover, when there are too many \textit{naive samples} in the training process, they can interfere with model performance improvement, as shown in Figure~\ref{fig:intro-a}. These findings highlight the importance of properly identifying and handling \textit{naive samples} in SSL tasks.

We would highlight that there are several factors that cause the slow performance improvement, such as the number of high confidence pseudo-labels, \etc.  
There are also multiple ways to solve the problem, such as adjusting the threshold~\cite{FlexMatch}, finding other learnable signals~\cite{simmatch}, \etc. In this work, we concentrate on data augmentation. It should be noted that our approach can be used together with the above ways and thus beneficial to SSL.

\section{SAA: Using Sample Adaptive Augmentation to Aid Semi-Supervised Learning}
SAA aims to address the issue of not effectively utilizing \textit{naive samples} by providing them with more attention and exploration of their value in aiding model training. To achieve this goal, SAA designs two modules: the sample selection module and the sample augmentation module. The function of the first module is to identify \textit{naive samples} in each epoch, while the function of the second module is to apply more diverse augmentation strategies to the \textit{naive samples} to facilitate their effective learning.

\subsection{Sample Selection}
We introduce two vectors $\mathcal{H}=\{\mathcal{H}_1,\mathcal{H}_2,...,\mathcal{H}_N\},\mathcal{F}=\{\mathcal{F}_1,\mathcal{F}_2,...,\mathcal{F}_N\}$,  where $N$ is the number of unlabeled samples.
$\mathcal{H}$ records historical consistency loss information for each unlabeled sample, and
$\mathcal{F}$ marks whether the unlabeled sample is a \textit{naive sample}.
For unlabeled sample $u_i$, the model calculates the consistency loss $l_i^t$ between its weakly- augmented version and strongly augmented versions once in $t$-th epoch. Then we update $\mathcal{H}_i^t$ with exponential moving average (EMA), which can be expressed as:
\begin{equation}
    \mathcal{H}_i^t = (1-a)\mathcal{H}_i^{t-1}+\alpha l_i^t.
\end{equation}

\begin{algorithm*}[]  
	\caption{Equip FixMatch with SAA}
	\LinesNumbered 
	\KwIn{Labeled data batch $\mathcal{B}_x=\{(x_i,y_i)\}^M$, unlabeled data batch $\mathcal{B}_u=\{u_i\}^N$, unsupervised loss weight $\lambda$, pre-training epochs $T'$, total training epochs $T$, augmentation strategies $\mathcal{\alpha}(\cdot), \mathcal{A}(\cdot), \mathcal{A'}(\cdot)$, historical loss $\mathcal{H}=\{\mathcal{H}_1,\mathcal{H}_2,...,\mathcal{H}_N\}$, mark $\mathcal{F}=\{\mathcal{F}_1,\mathcal{F}_2,...,\mathcal{F}_N\}$}
	\color{gray}
	\For{$t$ $\leftarrow 1$ to $T'$}{
	    Run FixMatch; \tcp{training as FixMatch in the first $T'$ epochs}
	}
	\For{\textcolor{gray}{$T'$ $\leftarrow 1$ to $T$}}{
		Compute $\mathcal{L}_{sup}={1}/{|\mathcal{B}_{x}|}\sum_{i=1}^{|\mathcal{B}_{x}|}H(q_i,y_i)$; \tcp{supervised loss in FixMatch}\   
		\For{$i$ $\leftarrow 1$ to $N$}{
		Apply perturbation $\mathcal{\alpha}(\cdot)$ to $u_i$; {}\tcp{weak augmentation in FixMatch}
		\textcolor{black}{Apply augmentation $\mathcal{A}(\cdot)$ / $\mathcal{A'}(\cdot)$ to $u_i$ according to $\mathcal{F}_i$; \textcolor[RGB]{120,152,225}{\tcp{{sample augmentation in SAA}}}}
		Compute loss $l_i=H(\arg \max p_m(\mathcal{\alpha}(u_i)), p_m(\mathcal{A}(u_i))) $; \tcp{consistency loss in FixMatch}
		}
		Compute $\mathcal{L}_{unsup}={1}/{|\mathcal{B}_{u}|}\sum_{i=1}^{|\mathcal{B}_{u}|}\mathbbm{1}(\max p_m(\mathcal{\alpha}(u_i))\ge\tau) l_i$; \tcp{unsupervised loss in FixMatch}
		\textcolor{black}{Update historical loss $\mathcal{H}_i^t = (1-a)\mathcal{H}_i^{t-1}+\alpha l_i^t$; \textcolor[RGB]{120,152,225}{\tcp{{sample selection in SAA}}}}
		\textcolor{black}{Update mark $\mathcal{F}_i=\mathbbm{1}(\mathcal{H}_i^t\le \texttt{OTSU}(H_i^t))$; \textcolor[RGB]{120,152,225}{\tcp{{sample selection in SAA}}}}
	}
\end{algorithm*}

Note that the parameter $\alpha$ introduced is not an additional model parameter, as the model parameters are also updated with EMA~\cite{FixMatch,FlexMatch}. Since the historical loss information can reflect the magnitude of the impact of a strongly augmented version on the model, it becomes the basis for our decision on \textit{naive sample}.
OTSU~\cite{otsu} is a commonly used method of threshold segmentation because of its computational simplicity, stability, and strong self-adaptation. Inspired by this, we calculate the historical loss threshold in each epoch:
\begin{equation}
    \tau_s = \texttt{OTSU}(\mathcal{H}_1,\mathcal{H}_2,...,\mathcal{H}_N).
\end{equation}
OTSU adaptively divides the sample into two parts based on the historical loss. Samples with small historical losses are considered as \textit{naive samples} since they provide less help to the model.
Then we update $\mathcal{F}$ by:
\begin{equation}
    \mathcal{F}_i=\mathbbm{1}(\mathcal{H}_i\le \tau_s).
\end{equation}

\begin{table*}[t]\scriptsize
    \centering
    \tabcolsep=4.25pt
    \begin{tabular*}{\hsize}{lllllllllll>{\kern-\tabcolsep}c<{\kern-\tabcolsep}|}
        \toprule
         &\multicolumn{3}{c}{CIFAR-10} &\multicolumn{3}{c}{CIFAR-100} &\multicolumn{3}{c}{SVHN}&\multicolumn{1}{c}{STL-10} \\
         Method&\multicolumn{1}{c}{40 labels}&\multicolumn{1}{c}{250 labels} &\multicolumn{1}{c}{4000 labels}& \multicolumn{1}{c}{400 labels}&\multicolumn{1}{c}{2500 labels} &\multicolumn{1}{c}{10000 labels}& \multicolumn{1}{c}{40 labels}&\multicolumn{1}{c}{250 labels} &\multicolumn{1}{c}{1000 labels} & \multicolumn{1}{c}{1000 labels}\\
         \cmidrule(r){1-1} \cmidrule(r){2-4}  \cmidrule(r){5-7} \cmidrule(r){8-10}\cmidrule(r){11-11}
         Mean-Teacher&      29.91$\pm$1.60 & 62.54$\pm$3.30 & 91.90$\pm$0.21 & 18.89$\pm$1.44 & 54.83$\pm$1.06 & 68.25$\pm$0.23 & 63.91$\pm$3.98 & 96.55$\pm$0.03 & 96.73$\pm$0.05 & -\\
         MixMatch&          63.81$\pm$6.48 & 86.37$\pm$0.59 & 93.34$\pm$0.26 & 32.41$\pm$0.66 & 60.42$\pm$0.48 & 72.22$\pm$0.29 & 69.40$\pm$8.39 & 95.44$\pm$0.32 & 96.31$\pm$0.37 & 38.02$\pm$8.29 \\
         ReMixMatch&        90.12$\pm$1.03 & 93.70$\pm$0.05 & 95.16$\pm$0.01 & 57.25$\pm$1.05 & 73.97$\pm$0.35 & \textbf{79.98$\pm$0.27} & 75.96$\pm$9.13 & 93.64$\pm$0.22 & 94.84$\pm$0.31 & 75.51$\pm$1.25 \\
         Dash&              91.84$\pm$4.31 & 95.22$\pm$0.12 & 95.76$\pm$0.06 & 55.17$\pm$1.36 & 72.15$\pm$0.19 & 77.23$\pm$0.21 & 96.97$\pm$1.59 & 97.83$\pm$0.10 & 97.97$\pm$0.06 & 83.17$\pm$0.80 \\
         CoMatch&           93.09$\pm$1.39 & 95.09$\pm$0.33 & 95.57$\pm$0.04 & - & - & - & - & - & - & 79.80$\pm$0.38\\ 
         SLA&               94.83$\pm$0.32 & 95.11$\pm$0.27 & 95.79$\pm$0.15 & 58.56$\pm$1.41 & 72.37$\pm$0.44 & 77.68$\pm$0.24 & 94.37$\pm$2.91 & 95.08$\pm$1.08 & 95.84$\pm$0.24 & -\\
         NP-Match&          95.09$\pm$0.04 & 95.04$\pm$0.06 & 95.89$\pm$0.02 & 61.09$\pm$0.99 & 73.97$\pm$0.26 & 78.78$\pm$0.13 & - & - & - & - \\
         SimMatch&          94.40$\pm$1.37 & 95.16$\pm$0.39 & 96.04$\pm$0.01 & \textbf{62.19$\pm$2.21} & 74.93$\pm$0.32 & 79.42$\pm$0.11 & - & - & - & 89.70$\pm$0.82\\
         \hline
         \specialrule{0em}{0.5pt}{1pt}
         FixMatch$^\dagger$ &92.50$\pm$0.67 & 95.10$\pm$0.04 & 95.81$\pm$0.05 & 53.17$\pm$0.51 & 72.64$\pm$0.17 & 77.60$\pm$0.09 & 96.24$\pm$0.98 & 97.54$\pm$0.04 & 97.98$\pm$0.02 & 85.27$\pm$1.15 \\
         \rowcolor{green!20}
         \textbf{FixMatch (w/SAA)}   &94.76$\pm$0.99 & 95.21$\pm$0.07 & 96.09$\pm$0.07 & 54.29$\pm$0.73 & 73.18$\pm$0.21 & 78.71$\pm$0.20 & \textbf{97.01$\pm$0.72}& \textbf{97.68$\pm$0.07} & \textbf{98.06$\pm$0.06} & 87.92$\pm$1.46 \\ 
         \hline
         \specialrule{0em}{0.5pt}{1pt}
         FlexMatch$^\dagger$&95.01$\pm$0.09 & 95.08$\pm$0.10 & 95.82$\pm$0.02 & 60.51$\pm$1.54 & 72.98$\pm$0.22 & 78.15$\pm$0.17 & 92.42$\pm$2.60 & 92.98$\pm$1.59 & 93.54$\pm$0.28 & 89.15$\pm$0.71 \\
         \rowcolor{green!20}
         \textbf{FlexMatch (w/SAA)} & \textbf{95.31$\pm$0.16} & \textbf{95.40$\pm$0.19} & \textbf{96.14$\pm$0.08} &61.87$\pm$1.94& \textbf{75.01$\pm$0.41} & {79.88$\pm$0.34}& 93.15$\pm$2.54 & 93.25$\pm$2.41 & 94.41$\pm$0.27 & \textbf{90.85$\pm$0.82}\\
         \hline
         \specialrule{0em}{0.5pt}{1pt}
         Fully-supervised & \multicolumn{3}{c}{95.38$\pm$0.05} &\multicolumn{3}{c}{81.70$\pm$0.09} &\multicolumn{3}{c}{97.87$\pm$0.02}&\multicolumn{1}{c}{-} \\
         \bottomrule
    \end{tabular*}
        \caption{Performance comparisons on CIFAR-10, CIFAR-100, SVHN, STL-10. We compare the performance with recent SSL works~\cite{MeanTeacher,ReMixMatch,MixMatch, Dash,CoMatch,SLA,simmatch,npmatch}. We apply SAA on the top of FixMatch~\cite{FixMatch} and FlexMatch~\cite{FlexMatch}, respectively. For fair comparison, we re-ran FixMatch and FlexMatch under the exact same random seed, which is denoted by $^\dagger$. Fully-supervised comparisons follows FlexMatch~\cite{FlexMatch}, which is conducted with all labeled date with applying weak data augmentations. Experiments shows SAA provides a significant improvement to the SSL model. When we choose FlexMatch as the base framework, performance reaches SOTA for most settings.}
    \label{tab:t1}
\end{table*}

It can be seen that the decision of the \textit{naive samples} is done at every epoch. In other words, whether a sample is a \textit{naive sample} is related to the model performance. We should note that there may be multiple shifts in $\mathcal{F}$ in the training process. On the one hand, if the sample is regarded as a \textit{naive sample}, we will apply a more diverse augmentation to it to avoid invalid learning. On the other hand, this more diverse augmentation may be too perturbing for the sample and negatively affect the model, so the augmentation strategy for these samples needs to be adjusted back to the original strategy in a timely manner.

\subsection{Sample Augmentation}
We apply different augmentations to the \textit{non-naive sample} and the \textit{naive sample}. The former will be applied with the original augmentation $\mathcal{A}$, while the latter will be applied with a new augmentation $\mathcal{A'}$, which increases the difficulty of the augmented version. This can be expressed as:
\begin{equation}
\texttt{Augmented}(u_i)=\left\{\begin{array}{l}
     \mathcal{A}(u_i),\mathcal{F}_i=1\\
     \mathcal{A'}(u_i),\mathcal{F}_i=0
\end{array}
\right.
\end{equation}

\begin{figure}[h]
    \centering
    \includegraphics[scale=0.7]{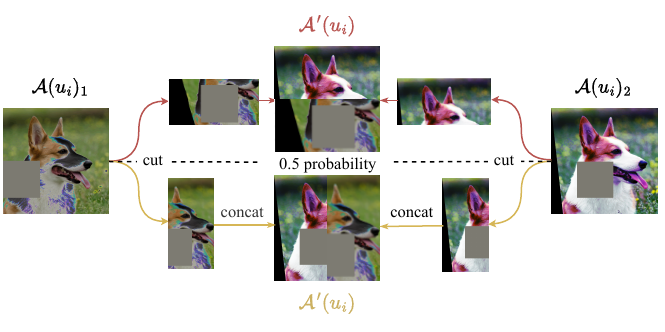}
    \caption{Augmentation for \textit{naive samples}. }
    \label{fig:method3}
\end{figure}

We implement this more diverse augmentation $\mathcal{A'}$ in a simple way, \ie, by regrouping several $\mathcal{A}(u_i)$ into a new image. 
Formally, a new augmented image $\mathcal{A'}(u_i)$ can be expressed as:
\begin{equation}
    \mathcal{A'}(u_i) = \texttt{Concat}(\texttt{Cut}(\mathcal{A}(u_i)_1), \texttt{Cut}(\mathcal{A}(u_i)_2)).
\end{equation}
As shown in Figure~\ref{fig:method3}, we have two strongly augmented images $\mathcal{A}(u_i)_1$ and $\mathcal{A}(u_i)_2$. To create a new augmented image $\mathcal{A'}(u_i)$, we randomly choose one of the following two options with equal probability: 1) Top-bottom concat: We take the top half of $\mathcal{A}(u_i)_1$ and the bottom half of $\mathcal{A}(u_i)_2$ and concatenate them to create a new image. 2) Left-right concat: We take the left half of $\mathcal{A}(u_i)_1$ and the right half of $\mathcal{A}(u_i)_2$ and concatenate them to create a new image.

Image regrouping is a technique to enhance image diversity by combining multiple augmented images into a new image. It is a simple and effective solution that has been shown to be effective in previous works such as "CutMix"~\cite{cut_once}. In comparison to learning-based data augmentation methods, which can also yield augmented images suitable for model learning, image regrouping has lower memory and computational overheads. However, in the case of CutMix, augmentation is done on the cut images, which may result in some loss of information about the original image. In contrast, in our method, augmentation is applied to the whole image, which preserves more information about the original image. This is discussed further in the experimental section of the paper.

\section{Experiments}

We used FixMatch and FlexMatch as a base framework to verify the validity of SAA on SSL benchmark datasets: CIFAR-10, CIFAR-100~\cite{cifar}, SVHN~\cite{svhn} and STL-10~\cite{STL}. In section~\ref{sec:ex1}, we present the specific implementation details. In section~\ref{sec:ex2}, we first verify that SAA can help the model improve test accuracy and achieve SOTA on SSL tasks. In addition, we compare the performance for the same number of iterations and verify that SAA can accelerate the model's improvement speed.

\subsection{Implementation Details}
\label{sec:ex1}
We adopt ``WideResNet-28-2"~\cite{WideResNet} for CIFAR-10 and SVHN, ``WideResNet-28-8"~\cite{WideResNet} for CIFAR-100 and ``ResNet18"~\cite{ResNet} for STL-10. For a fair comparison, we keep the same set of parameters as FixMatch and FlexMatch with $\{|\mathcal{B}_{\mathcal{X}}|=64,|\mathcal{B}_{\mathcal{U}}|=7|\mathcal{B}_{\mathcal{X}}|,\lambda=1\}$. The test model is updated by EMA with a decay rate of 0.999. $\mathcal{H}$ is updated in the same way and with the same parameters ($\alpha=0.999$). FixMatch and FlexMatch set the number of training iterations to $2^{20}$, and we keep this practice as well. In order to update the sample historical loss $\mathcal{H}$ and marker $\mathcal{F}$ in a timely manner, we consider every 1024 iterations as one epoch, \ie, a total of 1024 epochs are trained. As the augmentation $\mathcal{A}'$ is not suitable for the initial training of the model, we apply it only after the 100th epoch, while historical loss $\mathcal{H}$ is recorded from the beginning. We repeat the same experiment for five runs with different seeds to report the mean test accuracy and variance.

\begin{figure*}[t]
  \centering
  \begin{subfigure}{0.49\linewidth}
  \centering
    \includegraphics[scale=0.45]{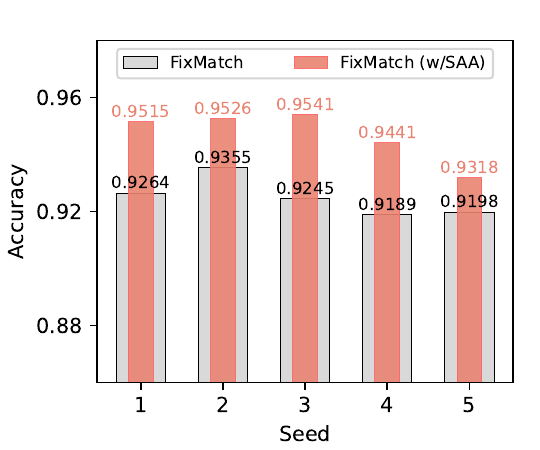}
    \includegraphics[scale=0.45]{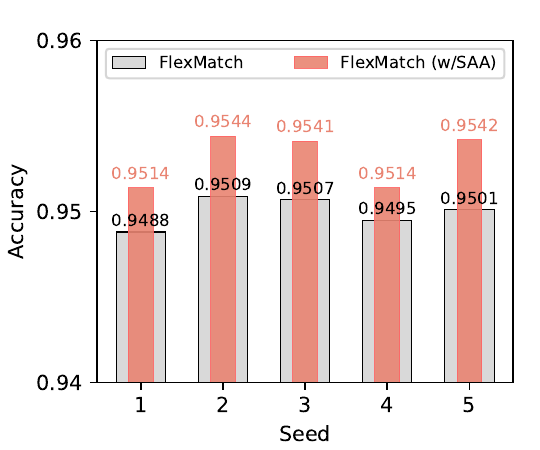}
    \caption{Accuracy under different seeds.}
    \label{fig:ex-1}
  \end{subfigure}
  \hfill
  \begin{subfigure}{0.49\linewidth}
  \centering
    \includegraphics[scale=0.45]{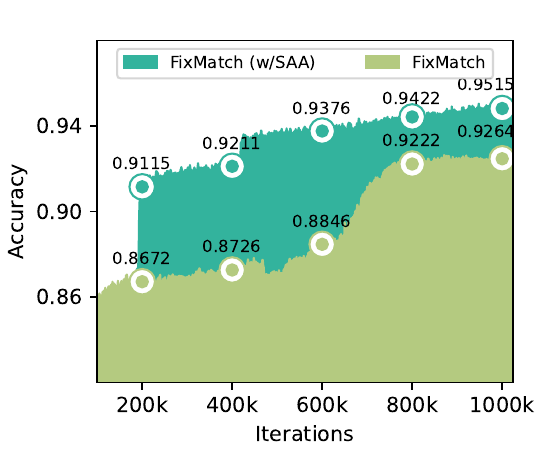}
    \includegraphics[scale=0.45]{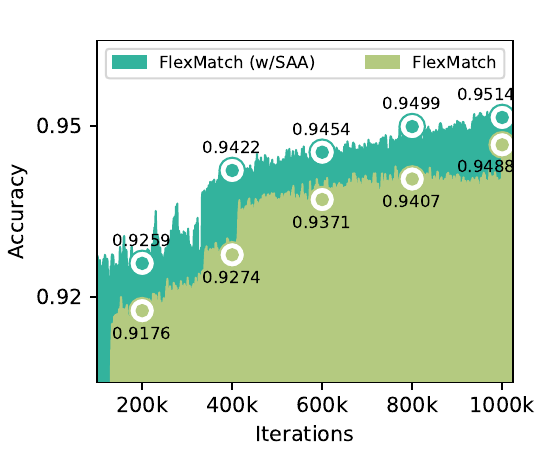}
    \caption{Performance during training (with the same seed).}
    \label{fig:ex-2}
  \end{subfigure}
  \caption{All experiments are conducted on task of CIFAR-10 with 40 labels. (a) shows the performance improvement of SAA on the model with 5 seeds. Although the magnitude of SAA's performance improvement on the model is different under different seeds, there is a steady improvement in general. (b) shows the performance growth during training. For the same iterations, SAA can significantly improve the performance of the model.}
  \label{fig:short}
\end{figure*}

\subsection{Main Results}
\label{sec:ex2}
\textbf{SAA improves the performance of baseline models.}  As shown in Tabel~\ref{tab:t1}, SAA successfully improves the test accuracy of FixMatch and FlexMatch under all settings. For instance, on CIFAR-10 with 40 labels, FixMatch and FlexMatch achieved a mean accuracy of 92.50\% and 95.01\%, while with SAA their average accuracy improved to 94.76\% and 95.31\%. For a  challenging and realistic task STL-10, SAA helped FixMatch to improve its accuracy by 2.65\% and FlexMatch by 1.70\%. FixMatch and FlexMatch outperform even under fully-supervised in some settings, \eg, FixMatch achieved mean test accuracy of 95.81\% on CIFAR-10 with 4000 labels and 97.98\% on SVHN with 1000 labels. We can notice that the variance of the model becomes slightly larger after applying SAA, as the boosting effect of SAA on the model is different under different seeds. As shown in Figure~\ref{fig:ex-1}, SAA boosts FixMatch at all 5 seeds, but it can boost by 2.51\% when the seed is 1 and 1.20\% when the seed is 2.

\textbf{We achieve the SOTA performance.} With FixMatch as the base, SAA can help to bring its performance up to near or even beyond that of other SSL models. For example, on CIFAR-10 with 40 labels, FixMatch (w/SAA) achieves an accuracy of 94.76\%, which is within 0.3\% of NP-Match. While on CIFAR-10 with 250 labels, FixMatch (w/SAA) achieves an accuracy of 95.21\%, which outperforms current SSL models. 
FlexMatch, which improves FixMatch by adjusting the threshold, can be further improved with the help of SAA. For example, on CIFAR-10 with 40 labels, SAA helped FlexMatch increase its mean accuracy to 95.31\%. For more difficult tasks, SAA helped FlexMatch increase its mean accuracy to 75.01\% and 90.85\% on CIFAR-100 with 2500 labels and STL-10, which outperforms all current SSL models.
Note that for unbalanced datasets, FlexMatch's threshold estimates for each class can produce large deviations, which is the reason for FlexMatch performs less favorably under the SVHN tasks. Since SVHN is a simple task, a fixed high threshold in FixMatch is more advantageous. When applying SAA to FixMatch, it is also possible to further improve its performance and outperform existing SSL models. 

\textbf{SAA accelerates the improvement of model performance.} Figure~\ref{fig:ex-2} shows the performance curve of the model training with the same seed. For example, at the 200k-th and 400k-th iterations, SAA helps FixMatch improve its performance from 86.72\% and 87.26\% to 91.15\% and 91.11\%, respectively. FlexMatch can also improve the performance of FixMatch for the same iterations by adjusting the confidence threshold so that the model can learn more samples. SAA, on the other hand, allows \textit{naive samples} to be learned more effectively, and therefore succeeds in further enhancing the learning of the model. For example, FixMatch reached an accuracy of 87.26\%, FlexMatch helped FixMatch improve to 92.74\%, and SAA helped FlexMatch further improve to 94.22\%. More often, we can observe that FixMatch encountered performance stagnant between approximately 200k-th and 600k-th iterations. This is because during this period the model learns a mass of strongly augmented versions that are non-useful for model performance improvement, while our proposed SAA successfully avoids this phenomenon by changing the augmentation for \textit{naive samples}.

\begin{table}[t]\scriptsize
  \centering
  \tabcolsep=4pt
  \begin{tabular}{@{}lcc@{}}
    \toprule
    \#Methods of selecting samples to apply $\mathcal{A}'$ & CIFAR-10 & STL-10   \\
    \midrule
    Baseline-1: Applying $\mathcal{A}$ to all samples & 92.50$\pm$0.67 & 85.27$\pm$1.15 \\
    Baseline-2: Applying $\mathcal{A}'$ to all samples & 92.98$\pm$2.94 &  83.19$\pm$3.98\\
    Baseline-3: Applying $\mathcal{A}'$ to 50\% samples (random) & 94.05$\pm$2.00 & 85.98$\pm$2.98\\
    \midrule
    Setting the threshold on $\mathcal{H}$: \\
    Fixed threshold (0.001) & 93.82$\pm$0.95 & 85.98$\pm$1.00\\
    Fixed threshold (0.002) & 94.10$\pm$1.22 & 85.22$\pm$1.87\\
    Fixed proportion threshold (25\%)& 93.10$\pm$0.89 & 85.38$\pm$1.20\\
    Fixed proportion threshold (50\%)& 93.87$\pm$1.52 & 86.08$\pm$1.97\\
    Fixed proportion threshold (75\%)& 93.85$\pm$2.29 & 84.29$\pm$2.03\\
    OTSU threshold & \textbf{94.50$\pm$1.05} & \textbf{87.92$\pm$1.46}\\
    \bottomrule
  \end{tabular}
  \caption{Different methods of selecting samples that applied with $\mathcal{A}'$. Experiment on conducted on the base of FixMatch. There are three ways to set the threshold on $\mathcal{H}$: 1) fixed value; 2) percentile of sorted $\mathcal{H}$; 3) automatic OTSU division.} 
  \label{tab:t3}
\end{table}

\section{Discussion}

\begin{figure*}[t]
  \centering
  \begin{subfigure}{0.49\linewidth}
  \centering
    \includegraphics[scale=0.45]{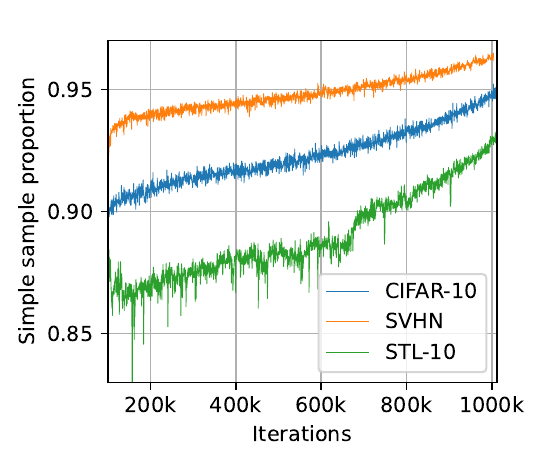}
    \includegraphics[scale=0.45]{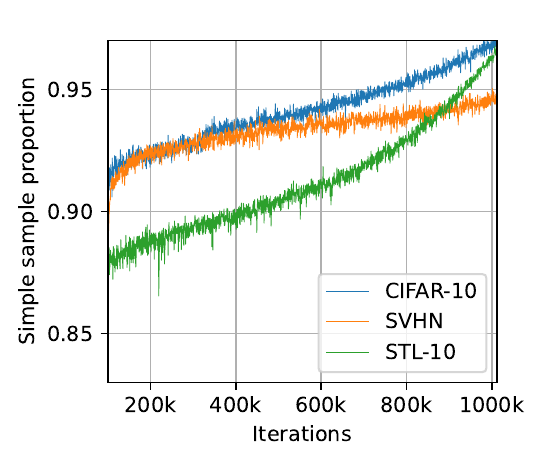}
    \caption{Proportion of \textit{naive sample}.}
    \label{fig:ex-3}
  \end{subfigure}
  \hfill
  \begin{subfigure}{0.49\linewidth}
  \centering
    \includegraphics[scale=0.45]{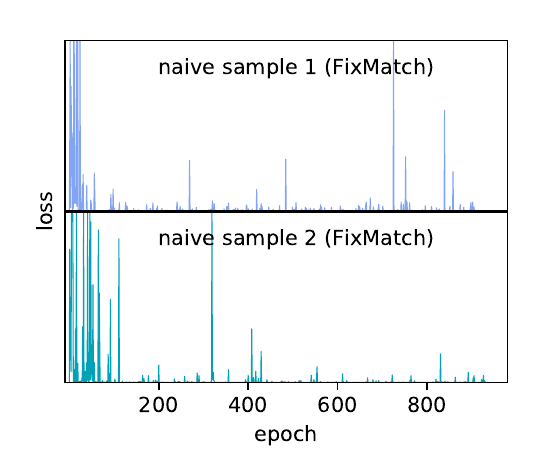}
    \includegraphics[scale=0.45]{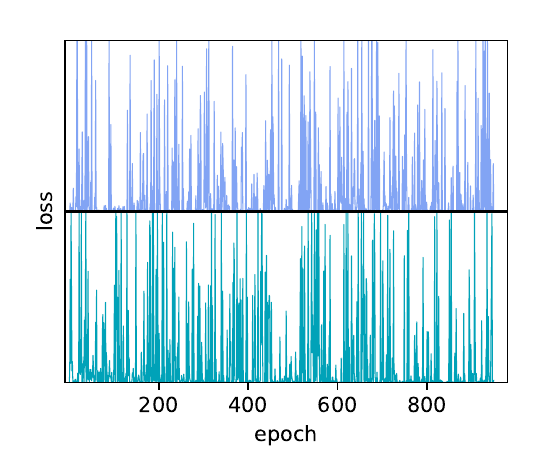}
    \caption{Consistency loss of two \textit{naive samples}.}
    \label{fig:ex-4}
  \end{subfigure}
  \vspace{-0.1cm}
  \caption{(a) shows the proportion of \textit{naive sample} divided by OTSU with model training. We recorded the proportion of \textit{naive samples} on CIFAR-10 with 40 labels, SVHN with 40 labels and STL-10. The pictures on the left and right are FixMatch (w/SAA) and FlexMatch (w/SAA) respectively. (b) shows the consistency loss of two \textit{naive samples}. The pictures on the left and right are FixMatch and FixMatch (w/SAA) respectively.}
  \label{fig:short}
  \vspace{-0.3cm}
\end{figure*}

\textbf{The more diverse augmentation is not applicable to all samples.} 
Our approach differs from~\cite{cut_once} in that we only apply the more diverse augmentation to a subset of samples (i.e., the \textit{naive samples}). 
We experimentally validated this, as shown in Table~\ref{tab:t3} with Baseline-1 and Baseline-2. 
It can be clearly seen that applying diverse augmentation to all samples can lead to instability and reduce performance on some tasks. 
This indicates that some images have too much semantic information corrupted under augmentation $\mathcal{A}'$, leading to an accumulation of errors. 
To further explore this, we applied $\mathcal{A}'$ to a randomly selected sample of 50\% at each epoch and the mean test accuracy of the model was slightly improved, but still unstable. This further gives us the sense that more diverse augmentation is necessary, but can only be used on the \textit{naive samples} to work better.

\textbf{Adaptively dividing \textit{naive samples}.} 
To identify the \textit{naive samples}, we use the historical consistency loss of the sample. We tested this approach on CIFAR-10 and STL-10 with different threshold settings. 
As shown in Table~\ref{tab:t3}, both the fixed threshold and fixed scale approaches have a boosting effect on the model, although the effect is unstable and varies for different datasets. 
For example, the fixed threshold $\tau_s=0.002$ outperforms better on CIFAR-10 task, while $\tau_s=0.001$ outperforms better on STL-10 task, so the fixed threshold will be a more tricky hyperparameter for different datasets. Compared with the first two approaches, OTSU is not only better adapted to cross-dataset tasks, but can also be tuned as the model is trained. Figure~\ref{fig:ex-3} shows the proportion of \textit{naive sample} divided by the OTSU method at different iterations. We can find that \textit{naive samples} are not only related to task difficulty, but also to model performance. For simpler datasets, the proportion of \textit{naive samples} is greater, and as model performance increases, more samples are also treated as \textit{naive samples}.

\begin{wrapfigure}{r}{3.5cm}
\vspace{-0.5cm}
\centering
\includegraphics[scale=0.4]{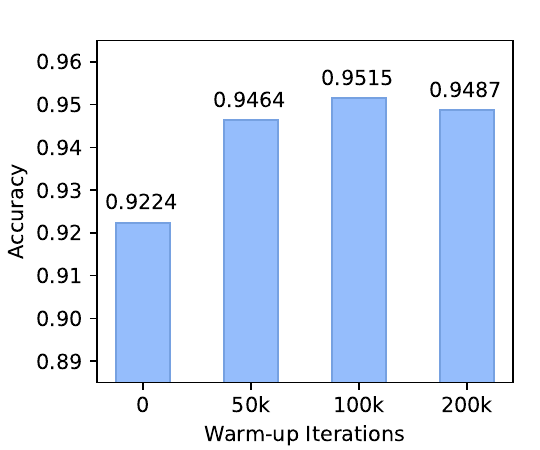}
\captionsetup{font=scriptsize}
\caption{Model warm-up. Experiments are conducted on CIFAR-10 with 40 labels.}
\label{fig:intro:twoprop}
\vspace{-0.3cm}
\label{fig:ex-5}
\end{wrapfigure}
\textbf{SAA prefers warm-up models.} We anaylze the model warm-up on of CIFAR-10 with 40 labels. The results in Figure~\ref{fig:ex-5} show that model warm-up with 100k iterations (10\% of total iterations) performs the best. This is because more diverse augmented images are more difficult to recognize, which can damage the initial training of the model. Therefore, SAA performs better on warm-up models. However, warming up the model too soon would reduce the action time of the SAA.

\textbf{SAA allows the augmented versions can further optimize the model.} 
Figure~\ref{fig:ex-4} compares the training loss of \textit{naive samples} with and without SAA in FixMatch. The plot shows that without SAA, the loss of \textit{naive samples} remains close to 0 most of the time after the 100th epoch, indicating that strongly augmented versions are not helpful for model training. However, with SAA, the number of times that strongly augmented versions aid model training significantly increases due to the dynamic adjustment of the augmentation strategy for these samples.

\begin{wrapfigure}{r}{3cm}
\vspace{-0.3cm}
\hspace{-0.75cm}
\centering
\includegraphics[scale=0.4]{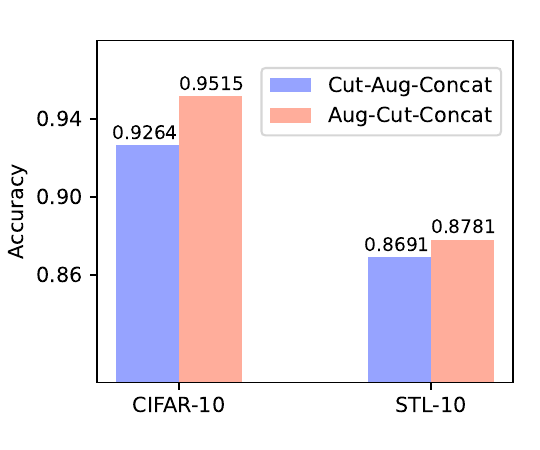}
\captionsetup{font=scriptsize}
\caption{Different more diverse augmentation. Experiments are conducted on CIFAR-10 with 40 labels and STL-10.}
\label{fig:intro:twoprop}
\vspace{-0.3cm}
\label{fig:ex-6}
\end{wrapfigure}\textbf{Augment on image, not on patches.} Previous work~\cite{cut_once} proposed to first cut an image into crops and then apply augmentation on them, while we instead apply augmentation on the whole image and then cut it into crops. We conducted experiments to compare these two methods, and the results are shown in Figure~\ref{fig:ex-6}. Our augmentation method performs better on both CIFAR-10 and STL-10 tasks. We attribute this to the fact that augmenting the whole image preserves more semantic information, which is safer for training SSL models.

\textbf{Limitations}. 
Since the augmentation we use is unlearnable, there is no guarantee that every augmented version is capable of contributing to model learning. Thus, the SAA serves to increase the likelihood that the augmented versions are useful to the model. In addition to this, if the model is already making good use of the sample, further augmentation may not be necessary or may even be detrimental to the model's performance, then the role of SAA is diminished.

\section{Conclusion}
In this paper, we first discuss the characteristics of \textit{naive samples} and their impact on model training and highlight that these samples should receive attention to uncover more value. We propose SAA to achieve this goal, which identifies \textit{naive samples} in real-time and dynamically adjusts their augmentation strategy so that they can contribute to model training. Our experimental results show that SAA significantly improves the performance of SSL methods, such as FixMatch and FlexMatch, on various datasets.

{\small
\bibliographystyle{ieee_fullname}
\bibliography{egbib}
}

\end{document}